\documentclass[11pt,a4paper]{article}
\usepackage{times}
\usepackage{url}
\usepackage{latexsym}
\usepackage{xspace}
\usepackage{color,soul}
\usepackage[defaultlines=4,all]{nowidow}

\usepackage[style=authoryear,maxbibnames=10,dashed=false,backend=bibtex]{biblatex}

\let \newcite \textcite
\let \cite \parencite

\bibliography{olid}

\DeclareRobustCommand{\eg}{\textit{e.g.}\@\xspace}
\DeclareRobustCommand{\ie}{\textit{i.e.}\@\xspace}

\makeatletter
\DeclareRobustCommand{\etc}{
    \@ifnextchar{.}%
        {etc}%
        {etc.\@\xspace}%
}
\makeatother

\DeclareRobustCommand{\ngs}{$n$-grams\@\xspace}

\newcommand\lidpy{\texttt{langid.py}\xspace}
\newcommand\nusvm{$\nu$-SVM\xspace}
\newcommand\ocsvm{OC-SVM\xspace}

\title{Open-Set Language Identification}

\author{Shervin Malmasi\\
	    Harvard Medical School\\
	    Boston, MA 02115, USA\\
	    {\tt smalmasi@bwh.harvard.edu}
}

\date{}

\begin{document}
\maketitle

\begin{abstract}
We present the first open-set language identification experiments using one-class classification models.
We first highlight the shortcomings of traditional feature extraction methods and propose a hashing-based feature vectorization approach as a solution.
Using a dataset of 10 languages from different writing systems, we train a One-Class Support Vector Machine using only a monolingual corpus for each language.
Each model is evaluated against a test set of data from all 10 languages and we achieve an average F-score of $0.99$, demonstrating the effectiveness of this approach for open-set language identification.
\end{abstract}

\section{Introduction}

Language Identification (LID) is the task of determining the language of a text, at the document, sub-document or even sentence level.
While LID is a fundamental preprocessing task in NLP, it is also widely used in information retrieval and filtering to select documents in a specific language; \eg LID can filter webpages or tweets by language.

LID is often framed as a multiclass text classification problem or tackled with statistical methods, which we outline in \S\ref{sec:olid-lidrelwork}.
Models are usually trained using data from a set of known languages.
However, a common criticism of these methods is that for any input, they select the closest matching class, even if the input is in an unknown language.

This is highly relevant as a very common LID use case is to detect whether a text is composed in a specific language;
other texts are often discarded.
In practice this means that training models for hundreds of languages and dealing with the dimensionality challenges may not be required.

Adding a new language requires re-training the full model, which can be slow.
However, this is a common scenario and especially relevant to the ever-increasing web/social media-based research.
Users often work with minority languages whose web-presence is growing as well as emerging dialects.
Example variants include script romanization (\eg romanized Russian or Persian) or Arabizi.
A recent study of African-American English is another salient example \cite{blodgett2016demographic}.

The motivation for this research is a practical one, driven by several scenarios such as the following example. A recent research project required the extraction of documents of a specific language from a large dataset.
In this case the particular language was Sorani, a Kurdish dialect that uses a writing system based on the Perso-Arabic script \cite{malmasi:2016:sorani}.
We tested a number of solutions, including the 
off-the-shelf \lidpy tool, none of which supported Sorani. Most of these classified our target documents as either Persian or Arabic, which use the same writing system. 

A good solution might be to re-train the models to include more languages. In this spirit, recent LID work has reported results on datasets including over $1{,}300$ languages \cite{brown2014non}, albeit using small samples.
However, considering the time, resources and computational costs of maintaining such systems, is this really the best solution for targeting a single language?

The main hindrance stems from the fact that most LID systems take a \textit{closed-set} recognition approach: they are trained on a closed set of classes and assume that all input belongs to one of these classes. This can be problematic when using such systems ``in the wild".
Researchers have noted that a different approach that allows categorization of all other unspecified languages is needed \cite{hughes2006reconsidering}.
Ideally we would be able to train such a method using only a monolingual corpus of our target language.

This is an issue that has been gaining wider recognition in recent years \cite{scheirer2013toward}, particularly in the field of computer vision.
\textit{Open-set classification} has recently been proposed an alternative approach to traditional closed-set classification.
This issue has also begun to attract attention within NLP \cite{fei2016breaking} and we believe that this will be an important research area in the coming years.

Accordingly, the aims of the present study are to (1) examine the issues revolving around feature extraction methods for open-set language identification
and
(2) evaluate the performance of one-class classification for language identification, training models using only monolingual data.

\section{Related Work}
\label{sec:olid-relwork}

\label{sec:olid-lidrelwork}
\subsection{Language Identification}
Work in language identification (LID) dates back to the seminal work of \newcite{Beesley:1988}, \newcite{dunning1994statistical} and \newcite{Cavnar:1994}.
Automatic LID methods have since been widely used in NLP research and applications.
Although LID has been widely studied, several open issues remain \cite{hughes2006reconsidering}. Current goals include developing models that can identify thousands of languages; extending the task to more fine-grained dialect identification; and making LID functionality available to users and developers.

LID research has also been extended to dialect identification \cite{malmasi:2015:lid,malmasi:2016:dslrec}, which aims to identify specific dialects of pluricentric languages such as Arabic \cite{malmasi-et-al:2015:adi}, German \cite{vardial2017}, and Persian \cite{malmasi-dras:2015:persian}.

In recent years LID has also been the focus of the ``Discriminating between Similar Language" (DSL) shared task series, which has also focused on more fine-grained dialect identification.
The 2015 edition attempted to move towards incorporating unknown languages by incorporating an additional ``other" class. This was an attempt to artificially create a negative class composed of a mixture of Catalan, Russian, Slovene, and Tagalog sentences \cite{zampieri:2015:LT4VarDial}.
Discriminative classifiers could easily distinguish between this class and the other targets and the it was not included in subsequent years.

Although it was easy to model this artificial ``other" class, this is not possible in the real world: while possible to obtain many positive examples, it is almost impossible to characterize every language or variant that is not a target.

\label{sec:olid-openset}
\subsection{Open-Set Recognition}

Open-set recognition has been gaining increased attention in recent years.
When used in the real world, classification systems will encounter data from previously unseen classes, also known as \textit{open space}.
Numerous methods, such as outlier and novelty detection, have been proposed to address this issue. While a detailed description is beyond the scope of this paper, the interested reader may refer to \newcite{scheirer2013toward} for details.

One-class classification (OCC) has been one successful approach to this problem.
Instead of identifying a decision boundary between the positive and negative class, OCC attempts to form a closed boundary that encapsulates the majority of the positive samples. All samples outside this boundary are considered outliers.
Unlike outlier and novelty detection methods, OCC is able to handle much larger feature spaces.
OCC has been successfully applied to image retrieval, acoustic scene classification \cite{battaglino2016open} and protein classification \cite{banhalmi2009one}

Within NLP, extending text categorization beyond the ``closed world" has continued to attract interest \cite{fei2016breaking}.
In this context, OCC also been used for document classification \cite{manevitz2001one} using One-Class Support Vector Machines (\ocsvm), one particular type of OCC based on the \nusvm model of \newcite{scholkopf2001estimating}.
SVMs are widely used for multi-class text classification \cite{malmasi:2015:mnli} and the \ocsvm
is a natural fit for text classification tasks.

The \ocsvm requires training data just from the positive class and only considers the origin as a negative example.
The $\nu$ parameter defines the upper-bound of outliers in the training data.\footnote{\eg $\nu=0.05$ assigns ~$95\%$ of the training data as inliers.}

\label{sec:olid-data}
\section{Data and Format}

Similar to the DSL task, we approach the problem at the sentence level, making it more challenging.
We needed data from various languages with distinct writing systems and compiled our dataset from several sources. The included languages and source corpora are shown in Table~\ref{tab:olid-data}.

\begin{table}
\centering
\begin{tabular}{|l|l|l|}
\hline
\bf Language & \bf Script & \bf Data Source(s) \\
\hline
Bulgarian	& Cyrillic & DSLCC, Europarl \\
Russian		& Cyrillic & DSLCC \\
\hline
Croatian	& Latin      & DSLCC \\
English		& Latin      & DSLCC \\
Spanish 	& Latin      & DSLCC \\
French		& Latin      & DSLCC \\
Slovak		& Latin      & DSLCC \\
\hline
Arabic		& Perso-Arabic & AOC, YouDACC \\
Persian		& Perso-Arabic & Collected \\
Kurdish		& Perso-Arabic & Collected \\
\hline
\end{tabular}
\caption{A listing of the languages in our dataset, their writing system and the source corpus. We collected 32k sentences per language.}
\label{tab:olid-data}
\end{table}

We use data from several sources, including the DSLCC corpus from the 2016 DSL task \cite{malmasi:2016:vardial}, the EuroParl corpus, the Arabic Online Commentary dataset at the Uppsala Persian Corpus. We supplied the Sorani Kurdish data from our own corpus.
For each of the 10 classes we collected $32{,}000$ sentences. Each of these languages will be used to train a single classifier that will be evaluated against all the other data.

\subsection{Data Format}

We encode and represent our data using the Unicode UTF-8 format. Although some researchers have approached this task at the byte level, we decided that a character-level representation would be suitable.
We also perform Unicode normalization. This is the process of converting Unicode strings so that all canonical-equivalent strings\footnote{
In Unicode, some sequences of code points may represent the same character: for example the character \"O can be represented by a single code point (U+00D6) or a sequence of the Latin capital letter O (U+004F) and a combining diaeresis (U+0308). Both will be rendered the same and are canonically equivalent, however, they will be processed as distinct features --- hence the need to perform normalization.}
have the exact same binary representation.\footnote{
More information can be found at \url{http://www.unicode.org/faq/normalization.html}}

\section{Features in Open Character Space}
\label{sec:olid-features}
Character \ngs are widely recognized as the best feature type for LID \cite{zampieri:2015:LT4VarDial} and they are the main feature we use here.
Traditional statistical methods extract features from the entire multi-class dataset and create a feature index or dictionary that will be used to create feature vectors. New features not seen during training will be discarded.

However, adapting this method to one-class training can be problematic: what do you do with unseen features? In our experiments we encountered several scenarios that we outline here.

\paragraph{No Overlap:} 
In the simplest scenario, a new sample will contain no previously seen features and can be confidently rejected as an outlier.

\paragraph{Lexical Borrowing and Noise:}
We observe that texts often include words from another language, often referencing entity names (people, books, \etc). Other noise such as URLs may also be present. 
If a one-class system trained on English is used to process a Kurdish string containing several English entity names, and the Kurdish \ngs are ignored as they were not previously seen, the text will be erroneously classified as English.

\paragraph{Writing System Overlap:}
The above problem is exacerbated when processing languages that have common letters, \eg English and French, as there will be many common \ngs.

Accordingly, we see that out-of-vocabulary features are informative and that the traditional feature extraction model cannot be used for this specific type of text classification task. We require a method that can preserve some of the information about these unseen features.

With respect to how a one-class model works (\S\ref{sec:olid-openset}), the unseen features would assist the model classify a sample as an outlier.
Ideally, our feature extraction method would be able to create a standard feature vector for any string, thereby allowing it to operate in an open character space.
This goal is facilitated by the fact that our system uses Unicode, which has both assigned and reserved character codepoints for new languages.

Recently, \textit{feature hashing}
has been proposed as a method to map an input to a feature vector by using a hash function \cite{weinberger2009feature}.
Hashing has proven to be simple, efficient and effective. It has been applied to various tasks including 
protein sequence classification \cite{caragea2012protein}, sentiment analysis \cite{da2014tweet}, and malware detection \cite{jang2011bitshred}.

This method uses a hash function $h(x)$ to arbitrarily map input to a hash key of a specified size.
The hash size, \eg $2^{18}$, determines the size of the mapped feature space.
Hash functions are many-to-one mappings. Collision occur when distinct inputs yield the same output, \ie $h(a) = h(b)$.
The collision rate is affected by the hash size. From a learning perspective, collisions cause random clustering of features and introduce noise;  unrelated features map to the same vector index and may degrade the learner's accuracy. However, it has been shown that ``the interference between independently hashed subspaces
is negligible with high probability" \cite{weinberger2009feature}.

A positive by-product of hashing is that it eliminates the need for a feature dictionary.
Consequently, a hashing-based feature vectorization method enables us to address the aforementioned challenges and vectorize any string, even if the characters are not included in our training data.
This is an ideal solution, and one which we employ in this study.

\section{Experimental Setup}

Having described our data (\S\ref{sec:olid-data}) and the feature extraction method (\S\ref{sec:olid-features}), we now describe our experimental setup.

We employ a linear-kernel \ocsvm model and set $\nu$ to $0.05$ as described in \S
\ref{sec:olid-openset}.
We train a single character 4-gram model for each of our languages.
For feature vectorization, our hash function is implemented using the signed 32-bit version of MurmurHash3 and we use a hash size of $2^{18}$.
Each model is trained on 90\% of the language data. The test set consists of the remaining 10\% of the positive instances ($3{,}200$ sentences) as well as all instances from the other $9$ classes ($288{,}000$ sentences).
In sum, each model is trained on $28{,}800$ sentences and tested on $291{,}200$ sentences from all 10 classes.

For evaluation, we measure per-class precision, recall and F1-score.
Given the large number of negative instances in the test sets, precision is a key metric for measuring performance while accuracy is not suitable due to the \textit{accuracy paradox}.

\section{Results}

We trained a model for each language and ran it against the test fold. The results for the 10 languages are presented in Table~\ref{tab:olid-results}. We observe almost perfect precision scores for all languages, and an average recall of $0.98$, resulting in an average F-score of $0.989$.
These results clearly demonstrate the effectiveness of a one-class model for language identification.

\begin{table}
\centering
\begin{tabular}{|r|l|l|l|}
\hline
\bf Language & \bf P & \bf R & $\mathbf{F_{1}}$ \\
\hline
Bulgarian	& $0.995$ & $0.980$ & $0.987$ \\
Russian		& $0.997$ & $0.985$ & $0.991$ \\
Croatian	& $1.000$ & $0.991$ & $0.996$ \\
English		& $0.998$ & $0.975$ & $0.986$ \\
French		& $1.000$ & $0.980$ & $0.990$ \\
Slovak		& $1.000$ & $0.982$ & $0.991$ \\
Spanish 	& $1.000$ & $0.980$ & $0.990$ \\
Arabic		& $1.000$ & $0.965$ & $0.982$ \\
Persian		& $1.000$ & $0.983$ & $0.991$ \\
Kurdish		& $1.000$ & $0.979$ & $0.989$ \\
\hline
\bf Average		& $1.000$ & $0.980$ & $0.989$ \\
\hline
\end{tabular}
\caption{One-class SVM classification results for each language using hashing-based feature vectorization based on character 4-grams. Each model is trained on 28k positive sentence, and tested against a set of $3{,}200$ sentences positive sentences as well as all sentence from the other $9$ classes ($288{,}000$ sentences).}
\label{tab:olid-results}
\end{table}

\section{Discussion and Conclusion}

We presented the first one-class classification experiments for open-set language identification, demonstrating the effectiveness of this methodology for the task.
We began by highlighting the shortcomings of traditional feature extractions methods used for multi-class classification. We proposed the use of a hashing-based feature vectorization method as an alternative, a method which has worked effectively for capturing and representing features from other languages.

Work in this area can lead to the creation of open-set language identification tools.
It is important to understand that the closed-set assumptions do not hold in the wild; most systems will encounter data from classes they were not trained on. It is important that instead of assigning them to the closest known class, the system should be sufficiently robust to reject these documents as coming from an unknown class.
The results described here hold promise for the development of
off-the-shelf tools for applying open-set language identification using just a monolingual corpus for training.
Such an approach will make it much easier for researchers to train models for their target linguistic varieties.

Open set classification is still a nascent topic in NLP and the present research is one of the first to apply this for LID.
This research is still at a preliminary stage, and more experiments with additional languages and parameters are needed to gain further insights about the methodology and its shortcomings.

There are a number of promising avenues for future work, much of which could not be included here due to space limitations.
Although our chosen \ocsvm model achieved good results, experiments comparing different model parameters or other one-class algorithms could also be insightful.
The extension of this work to include more languages, and possibly dialects, could be interesting. The application of ensemble architectures here could also boost results even further.

\interlinepenalty=10000
\printbibliography

\end{document}